\documentclass[pdflatex,sn-mathphys-num]{sn-jnl}

\usepackage{graphicx}%
\usepackage{multirow}%
\usepackage{amsmath,amssymb,amsfonts}%
\usepackage{amsthm}%
\usepackage{mathrsfs}%
\usepackage[title]{appendix}%
\usepackage{xcolor}%
\usepackage{textcomp}%
\usepackage{manyfoot}%
\usepackage{booktabs}%
\usepackage{algorithm}%
\usepackage{algorithmicx}%
\usepackage{algpseudocode}%
\usepackage{listings}%
\usepackage{subcaption}
\usepackage{adjustbox}

\usepackage{cellspace}

\theoremstyle{thmstyleone}%
\theoremstyle{thmstyletwo}%

\theoremstyle{thmstylethree}%

\definecolor{background}{HTML}{EEEEEE}
\definecolor{delim}{RGB}{20,105,176}
\colorlet{numb}{magenta}

\lstdefinelanguage{json}{
    basicstyle=\normalfont\ttfamily,
    numbers=left,
    numberstyle=\scriptsize,
    stepnumber=1,
    numbersep=8pt,
    showstringspaces=false,
    breaklines=true,
    frame=lines,
    backgroundcolor=\color{background},
    literate=
     *{0}{{{\color{numb}0}}}{1}
      {1}{{{\color{numb}1}}}{1}
      {2}{{{\color{numb}2}}}{1}
      {3}{{{\color{numb}3}}}{1}
      {4}{{{\color{numb}4}}}{1}
      {5}{{{\color{numb}5}}}{1}
      {6}{{{\color{numb}6}}}{1}
      {7}{{{\color{numb}7}}}{1}
      {8}{{{\color{numb}8}}}{1}
      {9}{{{\color{numb}9}}}{1}
      {:}{{{\color{delim}{:}}}}{1}
      {,}{{{\color{delim}{,}}}}{1}
      {\{}{{{\color{delim}{\{}}}}{1}
      {\}}{{{\color{delim}{\}}}}}{1}
      {[}{{{\color{delim}{[}}}}{1}
      {]}{{{\color{delim}{]}}}}{1},
}

\begin{document}

\title[PRISM]{PRISM: \textbf{P}atient \textbf{R}ecords \textbf{I}nterpretation for \textbf{S}emantic Clinical Trial \textbf{M}atching using Large Language Models }


\author[1]{\fnm{Shashi} \sur{Gupta}}\email{shashi.gupta@triomics.in}
\equalcont{These authors contributed equally to this work.}
\author[1]{\fnm{Aditya} \sur{Basu}}\email{aditya.basu@triomics.in}
\equalcont{These authors contributed equally to this work.}
\author[1]{\fnm{Mauro} \sur{Nievas}}\email{mauro.nievasoffidani@triomics.in}
\author[1]{\fnm{Jerrin} \sur{Thomas}}\email{jerrin.thomas@triomics.in}
\author[2]{\fnm{Nathan} \sur{Wolfrath}}\email{nwolfrath@mcw.edu}
\author[2]{\fnm{Adhitya} \sur{Ramamurthi}}\email{aramamurthi@mcw.edu}
\author[2]{\fnm{Bradley} \sur{Taylor}}\email{btaylor@mcw.edu}
\author*[2]{\fnm{Anai N.} \sur{Kothari}}\email{akothari@mcw.edu}
\author[1]{\fnm{Regina} \sur{Schwind}}\email{regina@triomics.com}
\author[3]{\fnm{Therica M.} \sur{Miller}}\email{therica.miller@mssm.edu}
\author[4]{\fnm{Sorena} \sur{Nadaf-Rahrov}}\email{sorena@ci4cc.org}

\author[5]{\fnm{Yanshan} \sur{Wang}}\email{yanshan.wang@pitt.edu}

\author*[1]{\fnm{Hrituraj} \sur{Singh}}\email{hrituraj@triomics.com}

\affil[1]{ \orgname{Triomics Research}, \city{San Francisco}, \country{USA}}

\affil[2]{ \orgname{Medical College of Wisconsin}, \city{Milwaukee}, \country{USA}}

\affil[3]{ \orgname{Icahn School of Medicine at Mount Sinai}, \city{New York}, \country{USA}}

\affil[4]{ \orgname{Cancer Informatics For Cancer Centers}, \city{Los Angeles}, \country{USA}}

\affil[5]{ \orgname{University of Pittsburgh}, \city{Pittsburgh}, \country{USA}}

\abstract{Clinical trial matching is the task of identifying trials for which patients may be potentially eligible. Typically, this task is labor-intensive and requires detailed verification of patient electronic health records (EHRs) against the stringent inclusion and exclusion criteria of clinical trials. This process is manual, time-intensive, and challenging to scale up, resulting in many patients missing out on potential therapeutic options. Recent advancements in Large Language Models (LLMs) have made automating patient-trial matching possible, as shown in multiple concurrent research studies. However, the current approaches are confined to constrained, often synthetic datasets that do not adequately mirror the complexities encountered in real-world medical data. In this study, we present the first, end-to-end large-scale empirical evaluation of clinical trial matching using real-world EHRs. Our study showcases the capability of LLMs to accurately match patients with appropriate clinical trials. We perform experiments with proprietary LLMs, including GPT-4 and GPT-3.5, as well as our custom fine-tuned model called OncoLLM and show that OncoLLM, despite its significantly smaller size, not only outperforms GPT-3.5 but also matches the performance of qualified medical doctors. All experiments were carried out on real-world EHRs that include clinical notes and available clinical trials from a single cancer center in the United States. 
}

\keywords{patient-trial matching, clinical trials, large language models, GPT4, automated matching, real-world datasets, medical data, inclusion criteria, exclusion criteria, therapeutic opportunities, empirical evaluation, custom-tuned models, medical doctors, patient records, research works, institute, efficacy, end-to-end, constraints, synthetic datasets}



\maketitle

\section{Introduction}\label{sec1}

Clinical trials are essential to advance scientific discovery, improve patient care and drive innovation in medicine\cite{Unger2016}. Their significance is particularly pronounced in oncology, where they can provide potential treatment options for patients with limited alternatives \cite{trialrecruitment1, trialrecruitment2, trialrecruitment3, trialrecruitment4}. Despite this importance, a substantial number of clinical trials encounter recruitment challenges. Only about 7\% of adults participate in cancer clinical trials \cite{Unger2016}. One of the major factors contributing to this recruitment bottleneck is the considerable challenge that physicians encounter when systematically reviewing each patient against the list of available trials.  This cumbersome task leads to a lower rate of trial recommendations \cite{lowrecommendation}, often due to the intricacy involved in deciphering the eligibility of a patient against the nuanced inclusion and exclusion criteria of these trials \cite{matching1, matching2, matching3, matching4, matching5}.

The core information pertinent to the inclusion and exclusion criteria of clinical trials are primarily found within unstructured EHRs, such as medical notes\cite{unstructured}. A recent study shows that this is particularly relevant for oncology, where key parameters for clinical trial screening are virtually absent in structured EHRs \cite{mcode}. These conditions create substantial challenges in interpretation at scale. Even when dealing with structured EHRs, creating queries using the trial criteria often proves challenging\cite{scaling, criteria2query}. To address this complexity, several research studies leveraging Natural Language Processing (NLP) have emerged, aiming to streamline the trial matching process. 

One common approach involves converting the inclusion and exclusion criteria of trials into structured queries\cite{elixr, electronicscreening, mcode, criteria2query}. These queries can then be utilized to efficiently retrieve relevant data from dedicated clinical data warehouses. An alternate strategy focuses on the extraction of key data elements from patient notes and reformatting this information into a structured layout to aid in filtering clinical trials. \cite{unstructured}. 

While both approaches have demonstrated effectiveness in clinical trial matching, they come with limitations. Both approaches frequently rely on rule-based engineering, which is cumbersome and inflexible. Additionally, scaling these approaches to encompass the diverse range of clinical trials and the variety of potential patient scenarios is also challenging. Recent advancements in leveraging LLMs for patient-trial matching show considerable promise\cite{trialgpt, scaling, Nievas2023DistillingLLMs, Wornow2024ZeroShotTrialMatching}. However, these studies predominantly utilize relatively simplistic or synthetic patient notes and clinical trial setups, which do not accurately reflect the complexity of real-world scenarios. Furthermore, they do not deal with the long context scenarios where a single patient may have hundreds of notes in their EHR. While \cite{scaling} attempts to deal with the entire patient journey, they restrict themselves to a well-defined variable while extracting them using an extensively trained model. Finally, most of these methods, if not all, rely on proprietary LLMs which are difficult to deploy in the sensitive and privacy focused healthcare domains \cite{genaiprobs}.

\begin{figure}
    \centering
    \includegraphics[width=1\linewidth]{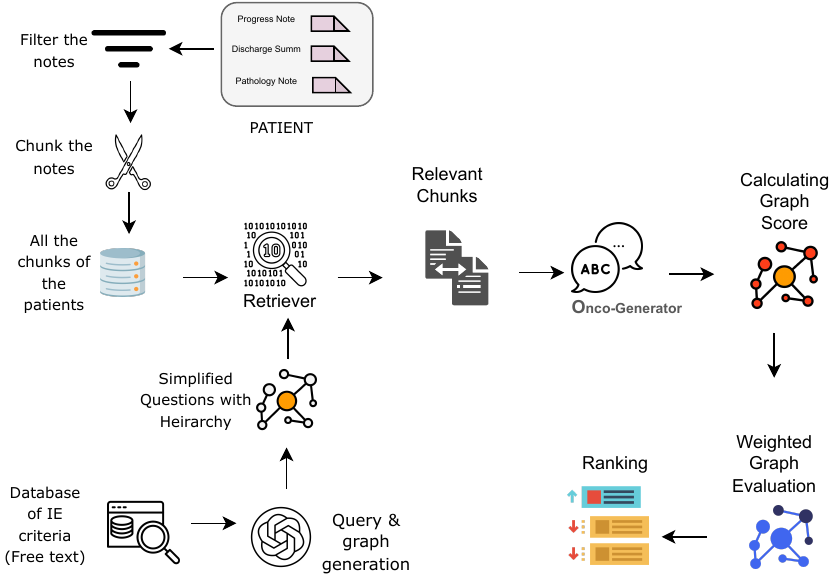}
    \caption{The pipeline only uses unstructured notes to effectively match the patients to potential clinical trials. Patient notes are first filtered as per the defined rules and are then chunked using a contextual chunker. The chunks are then stored in a database. The trial criteria are ingested as plain text as extracted from clinicaltrial.gov and are converted into a graphical question representation as described in Section \ref{sec:methods}. This graph is then used to retrieve relevant snippets of information, and our proprietary fine-tuned language model calculates a score for the graph. We then also apply weights to that graph using our developed heuristics, which allow the pipeline to rank the trials accurately.}
    \label{fig:enter-label}
\end{figure}

To the best of our knowledge, our study is the first to present a comprehensive pipeline to demonstrate an end-to-end clinical matching system using real-world patient EHR data and LLMs. This approach solely relies on the direct input of unstructured patient notes and the detailed clinical trial inclusion and exclusion criteria into the LLMs.

Our contributions can be summarized as follows:

\begin{enumerate}
  \item \textbf{Scalable End-to-End Pipeline}: We introduce a novel pipeline, PRISM, which performs patient records interpretation and directly uses the semantics of inclusion and exclusion criteria in clinical trials to match patients through a mechanism to directly ingest free text criteria into the pipeline without any rule-based processing.
  
  \item \textbf{Fine-tuned LLM}: Our custom-tuned model, OncoLLM, demonstrates superior performance over GPT-3.5 and comparable efficacy to GPT-4. OncoLLM is significantly smaller than both and can be hosted within private infrastructure to address privacy concerns.

  \item \textbf{Benchmarking Against Medical Doctors}: For the first time, we present evidence that LLMs can almost match the performance of qualified medical doctors for the task of clinical trial matching. This finding suggests the potential of LLMs for real-world clinical applications.

  \item \textbf{Comprehensive Evaluation}: We conduct an extensive evaluation of the proposed pipeline, not only in ranking clinical trials but also in providing detailed interpretation for the eligibility of patients with specific trial criteria.

  \item \textbf{Ranking Algorithm}: We propose a novel algorithm for ranking clinical trials, significantly improving the average position of relevant trials compared to a baseline approach.

  \item \textbf{Both Search Directionalities}: We demonstrate that the same pipeline used to identify trials for patients (patient-centric search) can also enable researchers to create eligible patient cohorts for a trial (trial-centric search). Our work is the first to demonstrate both directionalities using the same pipeline.
\end{enumerate}

\section{Related Work}
\subsection{Clinical Trial Matching}
The task of clinical trial matching, or patient-trial matching, has attracted significant attention from academia as well as industry. Based on the directionality of the search, patient-trial matching can either be patient-centered, i.e., identifying trials for a patient \cite{trialgpt, trecdataset, sigirdataset, reranking1, reranking2} or trial-centered, i.e., finding eligible patients given a trial \cite{cohortselection, criteria2query, parker, humanmachine}. Both have their individual importance with some overlaps. The patient-centered viewpoint offers a direct methodology for physicians to treat trials as a care option - an important goal of the cancer moonshot program\cite{mcode}. Trial-centric methodology, on the other hand, allows researchers to conduct pragmatic clinical trials or conduct feasibility analyses. In this study, we focus on patient-centered trial matching when we mention clinical trial matching, but we also show that our framework can be readily extended to support both directions.

One well studied approach involves converting the inclusion and exclusion criteria of a trial into a structured query and using that query to search over structured clinical data repository \cite{criteria2query, elixr, electronicscreening}. This, however, has limitations in oncology as the majority of information relevant for clinical trial screening for cancer patients is found in unstructured data \cite{mcode,scaling}. An alternative approach uses deep neural networks to directly convert patient information to an embedding representation to match against trials\cite{reranking1, reranking2}. Another method involves using patients already enrolled in trials as representations of those trials and employing embedding similarity techniques to find patients similar to those recruited\cite{casebased}. These techniques have, however, shown limited success in a complicated domain such as oncology.

\subsection{Large Language Models}
Language models have recently shown great promise for a variety of use cases in healthcare \cite{Singhal2023MedicalQA, Singhal2022LLMClinicalKnowledge, Nori2023FoundationModelsMedicine, Chen2023MEDITRON70B, need_clini}. For clinical trial matching, \textit{TrialGPT} \cite{trialgpt} demonstrated the capabilities of GPT-3.5\cite{Brown2020LanguageModels} in effectively ranking clinical trials for patients. This was succeeded by the \textit{TrialLlama} research \cite{Nievas2023DistillingLLMs}, which provided evidence that open-source models, specifically LLaMA 2\cite{Touvron2023Llama2}, could surpass GPT-3.5's performance, demonstrating their potential in privacy-sensitive applications. \cite{scaling} utilize GPT-4 but in a limited setup converting clinical trial criteria into a structured schema. Then they use Bidirectional Encoder Representations from Transformers (BERT) and alike models \cite {biobert, pubmedbert, sapbert, linkbert} to extract information from notes and use this structured information to check final eligibility. In parallel with our work, a study  by \cite{Wornow2024ZeroShotTrialMatching} employed GPT-4\cite{Achiam2023GPT4} and other open-source models for processing multiple patient notes while integrating retrievers to enhance cost-efficiency and overall effectiveness. However, their approach has a few significant limitations. It depended on a curated selection of patient charts containing only 2-5 notes, which is minimal compared to the usual 100-200 notes in real-world patient records. Furthermore, their method was tested on matching with a single clinical trial that used generic inclusion and exclusion criteria with only 13 criteria, greatly reducing the complexity typical of real-world clinical trials. In contrast, we evaluate the proposed pipeline in a setup with a number of potential trial candidates and the entire EHR of the patients, resulting in a dataset of over 200 trials and around 10,000 diverse clinical trial criteria. This substantial scope significantly differentiates our approach from previous studies.

\section{Results}\label{sec2}

\subsection{Criteria/Question Level Accuracy}
\begin{figure}[t]
  \centering
  \begin{subfigure}[b]{0.50\textwidth}
    \centering
    \begin{tabular}{|c|c|c|}
        \hline
        \textbf{}   & \textbf{} & \textbf{Without} \\
        \textbf{Model Name}   & \textbf{All} & \textbf{\textit{N/A}} \\
        \textbf{}   & \textbf{} & \textbf{samples} \\ \hline
        GPT3.5-Turbo          & 53\%           & 48\%                             \\ \hline
        Mistral-7B-Instruct   & 41\%           & 32\%                             \\ \hline
        Mixtral-8x7B-Instruct & 49\%           & 43\%                             \\ \hline
        Qwen14B-Chat          & 43\%           & 34\%                             \\ \hline
        \textbf{OncoLLM}      & \textbf{63\%}  & \textbf{66\%}                    \\ \hline
        GPT4                  & 68\%           & 72\%                             \\ \hline
        Expert Doctors*                 & 70\%           & -                               \\ \hline
    \end{tabular}
    \caption{\textbf{Question Level Accuracy}}
    \label{tab:cla}
  \end{subfigure}
  \hfill
  \begin{subfigure}[b]{0.42\textwidth}
    \centering
    \includegraphics[width=\textwidth]{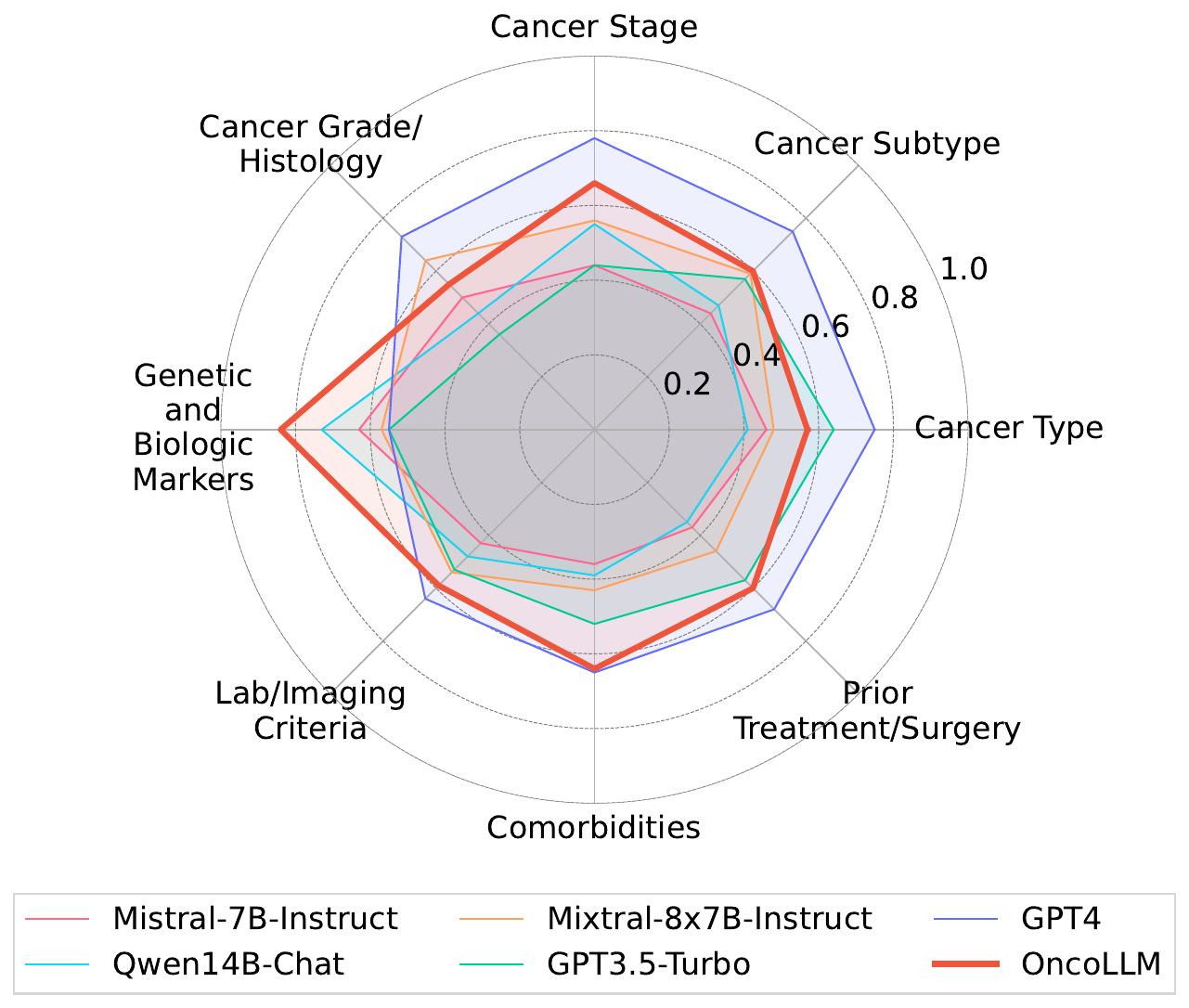}
    \caption{\textbf{Concept-Wise Accuracy}}
    \label{fig:fig-spider-cla}
  \end{subfigure}
  \caption{\textbf{(a) OncoLLM outperforms most of the prominent LLMs at criteria/question level answering accuracy.} First column \textbf{All} shows the question level accuracy across all the 720 Q\&A dataset for oncology related clinical trials. Second column \textbf{Without \textit{N/A} samples} shows question level accuracy after removing those questions whose answers were 'N/A' by medical experts. * Human accuracy was obtained only on 109 questions which was annotated by two medical experts. \textbf{(b) OncoLLM (in red) performs consistently well across all the relevant oncology related concepts.}}
  \label{fig:cla}
\end{figure}

\noindent\textbf{Accuracy Comparison:} We compare the accuracy of different LLMs in assessing whether a given patient meets a particular clinical trial's criteria. We evaluate the performance of OncoLLM model, a 14B parameter model, against several prominent LLMs, including Azure OpenAI's GPT-3.5-Turbo (175B), Qwen14B-Chat (14B)\cite{qwen}, Mistral-7B-Instruct (7B)\cite{jiang2023mistral}, Mixtral-8x7B-Instruct (56B)\cite{jiang2024mixture}, and Azure OpenAI's GPT-4. Table \ref{tab:cla} shows the performance of different LLMs at criteria/question level answering accuracy. TThe accuracy is defined as the percentage of instances in which the model's predictions match the clinically annotated answers for questions regarding patients, derived from the inclusion and exclusion criteria of clinical trials. Our results demonstrate that OncoLLM significantly outperforms both the similarly sized models and the larger GPT-3.5-Turbo, achieving an accuracy of \textbf{63\%} compared to 53\% by GPT-3.5-Turbo, 43\% by Qwen14B-Chat, 41\% by Mistral-7B-Instruct, and 49\% by Mixtral-8x7B-Instruct. Although GPT-4 reached a higher accuracy of 68\%, it incurs substantially greater computational costs, being approximately more than 100x times larger than OncoLLM (even though there are no official details on the parameter size of GPT-4, it is estimated to be more than a trillion parameters). Additionally, the Mistral-7B-Instruct model requires rule-based regex processing to output in JSON format, highlighting further limitations beyond raw accuracy. Another key aspect of this finding is that institutions can implement OncoLLM in their privacy-compliant infrastructure rather than transmitting sensitive patient data to external cloud servers. 

\begin{figure}[ht]
    \centering
    \includegraphics[width=1\linewidth]{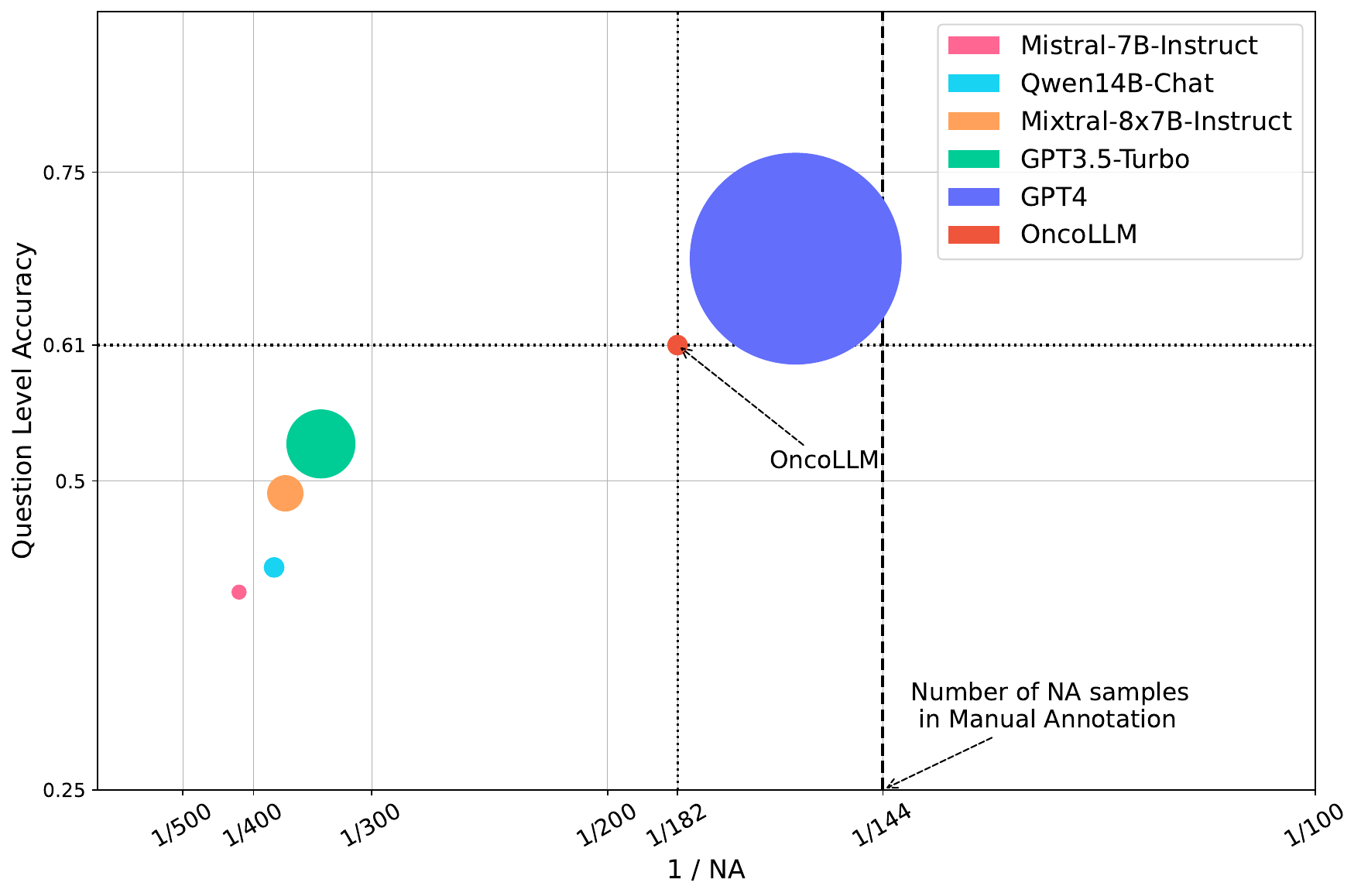}
    \caption{\textbf{Accuracy Comparison Based on Model Size and Number of "N/A" Outputs}. This figure presents a comparison of model accuracy with the frequency of "N/A" outputs. A higher frequency of "N/A" outputs indicates lower usefulness of the model. The size of each bubble represents the number of parameters of the model. This highlights the close performance of OncoLLM to GPT4 despite having relatively fewer parameters.}
    \label{fig:accuracy_comparison_bubble_graph}
\end{figure}

\noindent In an additional analysis where all questions initially marked as 'N/A' (meaning not enough information was available in patient records to answer those questions) were excluded to reduce ambiguity, we observed significant changes in model performances. By removing these uncertain inputs, the accuracy of our OncoLLM model increased to \textbf{66\%} (from 63\%), and the accuracy of Azure OpenAI's GPT-4 model also rose, reaching 72\% (from 68\%). Interestingly, the accuracy rates of Azure OpenAI's GPT-3.5-Turbo and the open-sourced models  decreased, suggesting that these models might rely more on ambiguous inputs to maintain higher performance levels or may exhibit less robustness in more clearly defined scenarios. This result indicates that some of the performance of the other 'weaker' models is inflated highlighting the gap between our model's performance and GPT-3.5. This is further illustrated in \textbf{Fig. \ref{fig:accuracy_comparison_bubble_graph}}.
\newline

\noindent\textbf{Concept-Level Scores:} We also conducted a comparison of the accuracy of each model on the Q\&A task at Concept level. For this analysis, we classified each question into various oncology-related clinical trial concepts, such as cancer type, cancer subtype, biomarkers, etc. These concepts were decided on the basis of physician input and were organized into different tiers based on their estimated clinical importance. Overall, we delineated 13 different concepts and categorized them into four tiers (\textbf{\textit{Tier 1, Tier 2, Tier 3, and Tier 4}}), with the importance level as \textbf{\textit{Tier 1 \textgreater{} Tier 2 \textgreater{} Tier 3 \textgreater{} Tier 4}} (See \textbf{Section S2 of Supplementary} for details).
We observed a similar pattern of accuracy at concept level as well.  OncoLLM generally secured the second position for most concepts. Notably, it outperformed GPT-4 in the biomarkers concept (\textbf{Fig. \ref{fig:fig-spider-cla}}).
\newline
\newline
\noindent\textbf{Annotation Process:} We manually annotated the Q\&A dataset from 10,000 notes and 50 cancer patients. We extracted 720 questions (each inclusion and exclusion criteria was converted into multiple questions with 'yes' or 'no' as possible answers), ensuring a balance across various patients, disease types, and categories of trial criteria (See \textbf{Section} \ref{sec:trial_composition}). To simplify the annotation, we used GPT-3.5 to identify and assess relevant sections for each question, reducing the potential workload from 800,000 segments to approximately 8,000. GPT-3.5 showed 98\% precision and 94\% recall in determining the relevance of segments to questions after refining our prompts. However, due to the high cost and token consumption ($>$ 2 billion) for processing, we opted not to use this method in our final setup as it would cost too much in the end to end pipeline for all the patients and trials. Five medical doctors annotated each question based on the chunks marked as relevant by GPT-3.5, labeling them as 'YES', 'NO', or 'N/A'. Each of the 720 questions was reviewed by at least one doctor, with about 12 questions reviewed by all five. The best performing annotators from this round participated in a second round, where 109 randomly selected questions were used to evaluate inter-annotator reliability. The average inter-annotator agreement (calculated as the percentage of times the annotators gave same answer to a particular question) among all five annotators was 64\% and 70\% for the two selected annotators (See \textbf{Section 1 of Supplementary}).
\newline
\newline

\subsection{Ranking Scores}

In this analysis, we conducted a comparative assessment of the ranking efficacy between our proposed approach (outlined in \textbf{Section \ref{sec:methods}}) for OncoLLM compared to GPT-3.5-Turbo model. Due to the cost associated with matching 1000 patient-trial pairs, we did not evaluate GPT4 for this task. To facilitate ranking, we utilized a scoring module (detailed in \textbf{Section \ref{sec:scoring_module}}), which assigns a matching score to each patient-trial pair. This score was used to rank trials for a patient and vice-versa.
\newline

\begin{figure}[t]
    \centering
    \begin{subfigure}[b]{0.49\textwidth}
        \centering
        \includegraphics[width=0.99\linewidth]{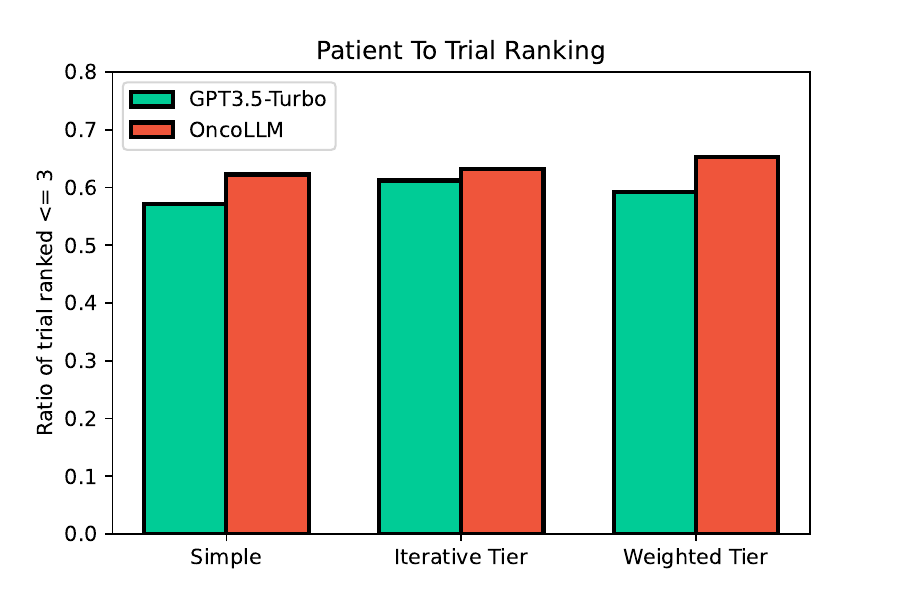}
        \caption{The proportion of patients for whom their respective ground truth trial ranked within the top-3. Here, "ground truth trial" is the trial in which the patient was enrolled.}
        \label{fig:patient_to_trial_ranking}
    \end{subfigure}
    \hfill
    \begin{subfigure}[b]{0.49\textwidth}
        \centering
        \includegraphics[width=0.99\linewidth]{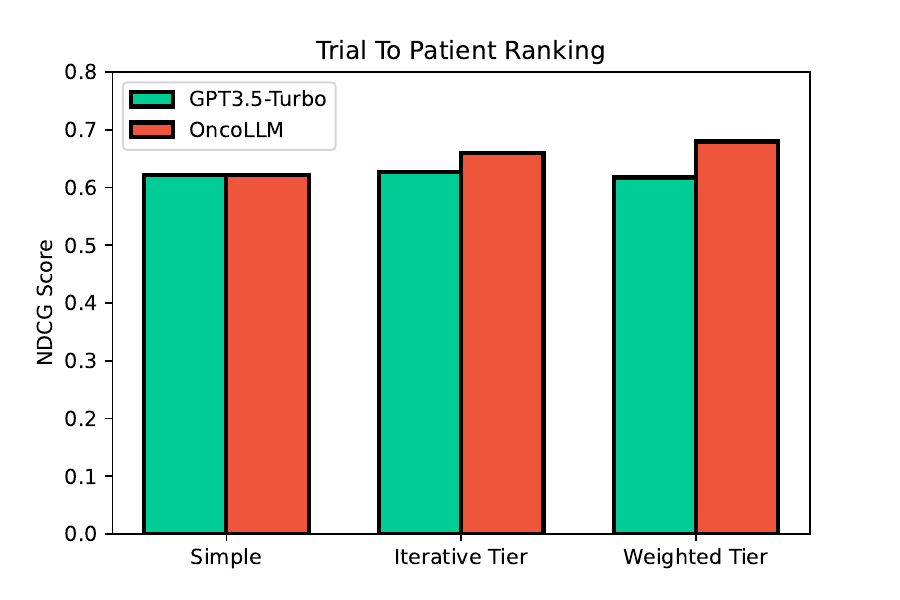}
        \caption{NDCG score for ranking patients within a specified trial. Patients who were initially enrolled for the specified trial are assigned a relevance score of 1, while rest are assigned 0.}
        \label{fig:trial_to_patient_ranking}
    \end{subfigure}
    \caption{\textbf{OncoLLM with \textit{Weighted Tier} scoring method performs best for both way search.} \textbf{(a)} OncoLLM (\textit{Weighted Tier}) ranked ground truth trials \textbf{65.3\%} of times in the top-3 among 10 considered trials, while GPT3.5-Turbo (\textit{Iterative Tier}) ranked ground truth trials only 61.2\% of times in the top-3. \textbf{(b)} OncoLLM (\textit{Weighted Tier}) scored an NDCG score of \textbf{68\%} as compared to 62.6\% of GPT3.5-Turbo (\textit{Iterative Tier}). See \textbf{Section \ref{sec:scoring_module}} for details on the scoring methods.}
\end{figure}

\subsubsection{Patient-Centric Ranking}\label{sec:patient-trial-matching}

\noindent\textbf{Evaluation Metric:} For this evaluation, we assembled a cohort of 98 cancer patients who had participated in clinical trials. We identified 10 real-world trials for each patient, all of which shared the same cancer disease type (e.g., lung cancer, breast cancer, etc.) and were actively recruiting patients at the time the patient enrolled in a clinical trial. Among these 10 trials, one trial served as the ground truth trial, denoting the trial in which the patient was actually enrolled (see section \ref{sec:dataset_preparation}). To assess the performance, we analyzed the proportion of times the rank of the ground truth trial fell within the top-3 ranks. This metric directly reflects the practical utility of our method, as it aids in the efficient shortlisting of eligible trials for patients within a hospital or institution setting.
\newline

\noindent\textbf{Analysis:} In our assessment, OncoLLM demonstrated superior performance compared to GPT-3.5-Turbo across all three scoring methods. Specifically, with the \textbf{\textit{Weighted Tier}} method, OncoLLM achieved a score of \textbf{0.65}, outperforming GPT-3.5-Turbo's score of 0.59. Similarly, with the \textbf{\textit{Iterative Tier}} method, OncoLLM achieved a score of \textbf{0.63}, surpassing GPT-3.5-Turbo's score of 0.61. Additionally, with the \textbf{\textit{Simple}} method, OncoLLM attained a score of \textbf{0.62} compared to GPT-3.5-Turbo's score of 0.57. These findings are consistent with our observations regarding question-level accuracy (\textbf{Fig. \ref{tab:cla}}); typically, a model exhibiting higher accuracy in answering eligibility questions tends to perform better in ranking tasks.
\newline

\begin{figure}
    \centering
    \includegraphics[width=1\linewidth]{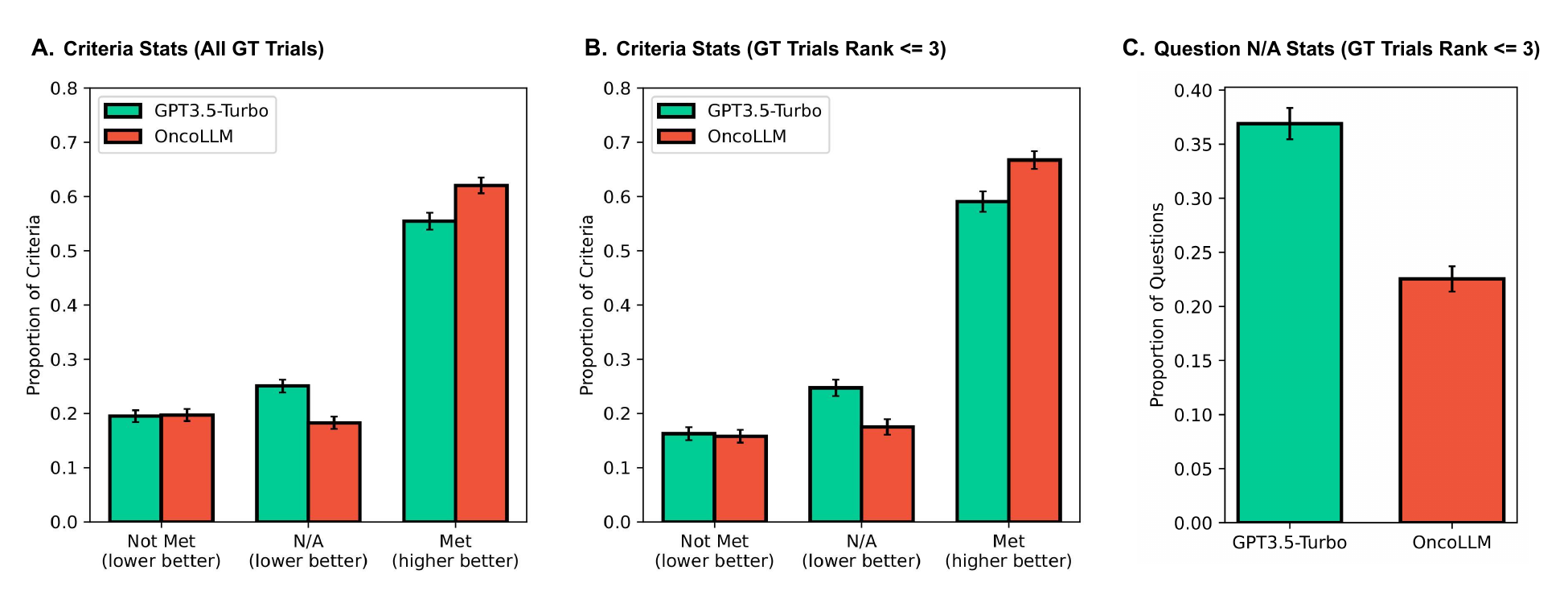}
    \caption{\textbf{Criteria/Question Level Analysis on 98 Patient, Ground Truth Trial Pairs}. \textbf{A.} Criteria level Met/Not-Met/NA stats for all the ground truth trials. \textbf{B.} Criteria level Met/Not-Met/NA stats where the ground truth trial ranked within the top-3. \textbf{C.} Question level N/A stats where the ground truth trial ranked within the top-3 (lower is better).}

    \label{fig:gt_trial_pt_matching_stats}
\end{figure}

\noindent\textbf{Criteria/Question Level Analysis:} To evaluate the utility of both GPT-3.5-Turbo and OncoLLM, we conducted an additional analysis for all patient and ground truth trials. Ideally, for ground truth trials where a patient is already enrolled, all eligibility criteria should be met. With this hypothesis in mind, we examined the statistics of all eligibility criteria for the ground truth trials. A superior system should meet a higher number of criteria, and produce fewer indecisive cases (instances where the system failed to produce a definitive Met/Not Met result for the given criteria, referred to as 'N/A' henceforth). We discovered that OncoLLM meets 62\% of criteria on average across all 98 patients, compared to GPT3.5-Turbo's score of 55.4\% (\textbf{Fig. \ref{fig:gt_trial_pt_matching_stats}}A). This analysis was extended to instances where the ground truth trial ranked within the top-3. Predictably, the overall criteria Met numbers increased for both models, with OncoLLM (66.7\%) outperforming GPT3.5-Turbo (59\%). Intriguingly, the overall average for OncoLLM across all ground truth trials remains higher than GPT3.5-Turbo's score for top-3 ranked ground truth trials. When it comes to ranking trials, OncoLLM's outcomes prove more practical as GPT3.5-Turbo tends to return 'N/A' responses to simplified questions derived from top-3 ranked ground truth trials considerably more often than OncoLLM (\textbf{Fig. \ref{fig:gt_trial_pt_matching_stats}C}, \textbf{Fig. \ref{fig:accuracy_comparison_bubble_graph}}). When a human is involved in the loop, this becomes problematic as N/A responses provide very little information to determine the patient's final eligibility.

\subsubsection{Trial-Centric Ranking}

\noindent\textbf{Evaluation Metric:} For this evaluation, we assembled a set of 36 clinical trials, all of which recruited patients from the same institution. For each clinical trial, we compiled a group of patients who shared the same cancer disease type and were enrolled in some trial when the specified clinical trial was active. This patient set also included those who were actually enrolled in the specified trial. On average, each trial had $2 \pm 1$ ground truth patients and $13 \pm 8$ additional patients. Here, 'ground truth patients' refer to those who were enrolled in the specified trial. Based on this data, we computed the NDCG score for ranking patients for a given clinical trial. To calculate the NDCG score, we utilized a binary relevance score, where ground truth patients were assigned a relevance score of 1, and the remaining patients were assigned a relevance score of 0.
\newline

\noindent\textbf{Analysis:} In our evaluation, OncoLLM exhibited superior performance compared to GPT3.5-Turbo across all scoring methods, except for the \textbf{\textit{Simple}} method where both models achieved similar scores. Specifically, using the \textbf{\textit{Weighted Tier}} method, OncoLLM attained a score of 0.68, outperforming GPT3.5-Turbo's score of 0.62. Similarly, with the \textbf{\textit{Iterative Tier}} method, OncoLLM achieved a score of 0.66, surpassing GPT3.5-Turbo's score of 0.63. For the \textbf{\textit{Simple}} method, both models scored 0.62. Once again, these results align with our observations regarding question-level accuracy (\textbf{Fig. \ref{tab:cla}}); typically, a model demonstrating higher accuracy in answering eligibility questions should excel in ranking tasks.
\newline
\newline

\subsection{Error Analysis}

\noindent\textbf{Trial Level:} While the patients were enrolled in a specific trial, this did not preclude their eligibility for other concurrent trials. Consequently, the ranking numbers do not accurately reflect the model's performance strength. To further assess this, we engaged a qualified clinical data expert to manually check the eligibility of patients for 10 randomly chosen trials that OncoLLM ranked in the top-1, but which were not the trials in which the patients were actually enrolled. We discovered that out of these 10 trials, the patients were deemed eligible for 9, effectively increasing the actual top-3 accuracy to $\approx95\%$. Interestingly, the only trial for which a patient was ineligible was due to the presence of tumors in both breasts; the patient was eligible for the trial concerning the tumor in the left breast but not for the one in the right breast. This highlights OncoLLM's effectiveness in identifying suitable trials for patients.
\newline

\noindent\textbf{Criteria Level:} To assess the capability of OncoLLM in generating accurate interpretation and citations of corresponding trial criterion to a patient, a set of questions (provided in supplement) and their corresponding responses generated by the model were randomly selected and reviewed by qualified medical professionals. These reviewers were tasked with verifying the correctness of the responses (categorized as "Yes", "No", or "N/A") and the accuracy of interpretation and citations provided with each answer. Our findings indicated that the accuracy of the final answers was \textbf{75.26\%}. When the final answer was deemed correct, the accompanying explanations were accurate \textbf{90.91\%} of the time. Additionally, citations included in the explanations were rated as correct \textbf{86.71\%} of the time and partially correct \textbf{6.29\%} of the time, highlighting OncoLLM's effectiveness in delivering reliable information and its utility in supporting further manual verification processes.

\subsection{Cost-Benefit Analysis}
We also conducted cost analysis to estimate the expenses associated with running different OncoLLM vs Azure OpenAI's GPT-4 for patient-trial matching, considering 98 patients with 10 trials for each patient (refer to \textbf{Section \ref{sec:patient-trial-matching}}). For estimation, we calculated the input prompt token and expected generated token count, from which we derived the pricing. For OncoLLM, we benchmarked the input and output generation speed (tokens/sec) when hosted using vLLM and calculated the running time using the formula:

\begin{equation}
    running\ time = (\frac{\# \ of\ input\ tokens}{input\ speed*3600} + \frac{\# \ of\ output\ tokens}{output\ speed*3600}) \text{   hours}
\end{equation}
\newline
\\
The final cost is determined by multiplying the hourly cost of Google Cloud GPU VM by the running time. For Azure OpenAI's GPT-4 model, we estimated the price using their \href {https://azure.microsoft.com/en-in/pricing/details/cognitive-services/openai-service/} {pricing table}.
\newline
\\
Our calculations indicate that the cost of operating OncoLLM is approximately \$170, while running GPT-4 incurs an expense of around \$6055. This implies an expense increase of about 35-fold. When we dissect the cost to compute the expense of a single patient-trial match, OncoLLM costs approximately \$0.17 per patient-trial pair, whereas GPT-4 demands \$6.18 per patient-trial pair. In addition, we have not considered several other optimizations available for LLMs \cite{FlashAttention-2, TensorRT-LLM} in this assessment. Other factors not taken into account include the availability of various cloud services offering more affordable computation, as well as the potential for a local in-house setup.
\newline

\section{Methods}\label{sec:methods}
\subsection{Problem Formulation}
We formulate the problem of matching patients to clinical trials through a compositional question-answering framework. Consider a patient \( P \), associated with a set of patient notes \( N = \{n_1, n_2, \ldots, n_i\} \), where each \( n_i \) contains distinct aspects of the patient's medical history. Each clinical trial is defined by its criteria text \( T \). From \( T \), a set of questions \( Q = \{q_1, q_2, \ldots, q_j\} \) is systematically derived, where each \( q_j \) corresponds directly to specific inclusion or exclusion criteria.

The set of possible answers to each question, denoted as \( A = \{a_1, a_2, \ldots, a_j\} \), is extracted through our pipeline from the dataset \( N \). Each element in \( A \) corresponds to an answer which can be 'Yes', 'No', or 'NA'. Here, 'NA' indicates that the question is not applicable to the patient in question or that there is insufficient information available to provide a conclusive answer. Conversely, 'Yes' and 'No' explicitly indicate whether the patient meets or does not meet the specified criteria, respectively.

A logical composition function \( C \) is then employed to integrate these answers into a single framework. This function \( C \) is defined by logical operators and possibly fuzzy logic controls that aggregate the individual answers based on the logical interdependencies specified in \( T \). The output of this function is a composite score \( S \), calculated as \( S = f(C(a_1, a_2, \ldots, a_j)) \), where \( f \) is a scoring function that quantitatively evaluates the overall compliance of the patient data \( N \) with the trial criteria \( T \).

The score \( S \) thus represents a quantified suitability of the clinical trial for patient \( P \), enabling the ranking of potential trials. Higher values of \( S \) indicate a better match, facilitating prioritization of trials for which the patient is most eligible. 

\subsection{Dataset Preparation}\label{sec:dataset_preparation}
Patients were identified using an institutional clinical research data warehouse that includes deidentified clinical notes and clinical trial enrollment information. To develop the analytic dataset, we selected patient note types that include oncology-relevant information (See \textbf{Section S3 of Supplementary}).
For each patient \( P \), who may have been enrolled in one or multiple trials \( T_1, T_2, \ldots, T_n \), we selected a single trial based on the complexity of its criteria, as determined by the length of the text in the inclusion and exclusion sections. This process resulted in identifying one 'ground truth' trial for each patient. The patient notes were limited to include information only until the enrollment date of the patient.

This curated dataset comprised 98 patient-trial combinations. To facilitate the ranking experiments, it was necessary to identify potential negative trials for each patient. We achieved this by first filtering both patients and trials based on the type of cancer. We then searched the trial database and retained a trial in a patient's set of negatives if it satisfied all of the following criteria: (1) the patient was not previously enrolled in that trial, and (2) the trial was active according to ClinicalTrials.gov at the time of the patient's enrollment. This method yielded 980 patient-trial combinations.

\begin{figure}[t]
    \centering
    \begin{tabular}{|l|c|c|}
        \hline
        \textbf{Category} & \textbf{\# of Patients} & \textbf{Average \# of Documents} \\
        \hline
        Breast Cancer     & 37                  & 49                           \\
        \hline
        Lung Cancer       & 20                  & 76                           \\
        \hline
        Prostrate Cancer  & 29                  & 57                           \\
        \hline
        Colorectal Cancer & 7                   & 283                          \\
        \hline
        Skin Cancer       & 4                   & 52                           \\
        \hline
    \end{tabular}
    \caption{Distribution of Patients by Cancer Type}
    \label{tab:cancer_data}

\end{figure}

\subsection{PRISM Pipeline}

Our end-to-end PRISM pipeline comprises several modules as shown in Fig \ref{fig:enter-label}. It takes clinical trial information and patient notes as input and provides a scalar score for the match between patient and trial. Once we have the score, it is used to rank patients by calculating the score of that trial for multiple patients or vice-versa.
\newline
\subsubsection{Trial Composition Module}\label{sec:trial_composition}
\begin{figure}
    \centering
    \includegraphics[width=1\linewidth]{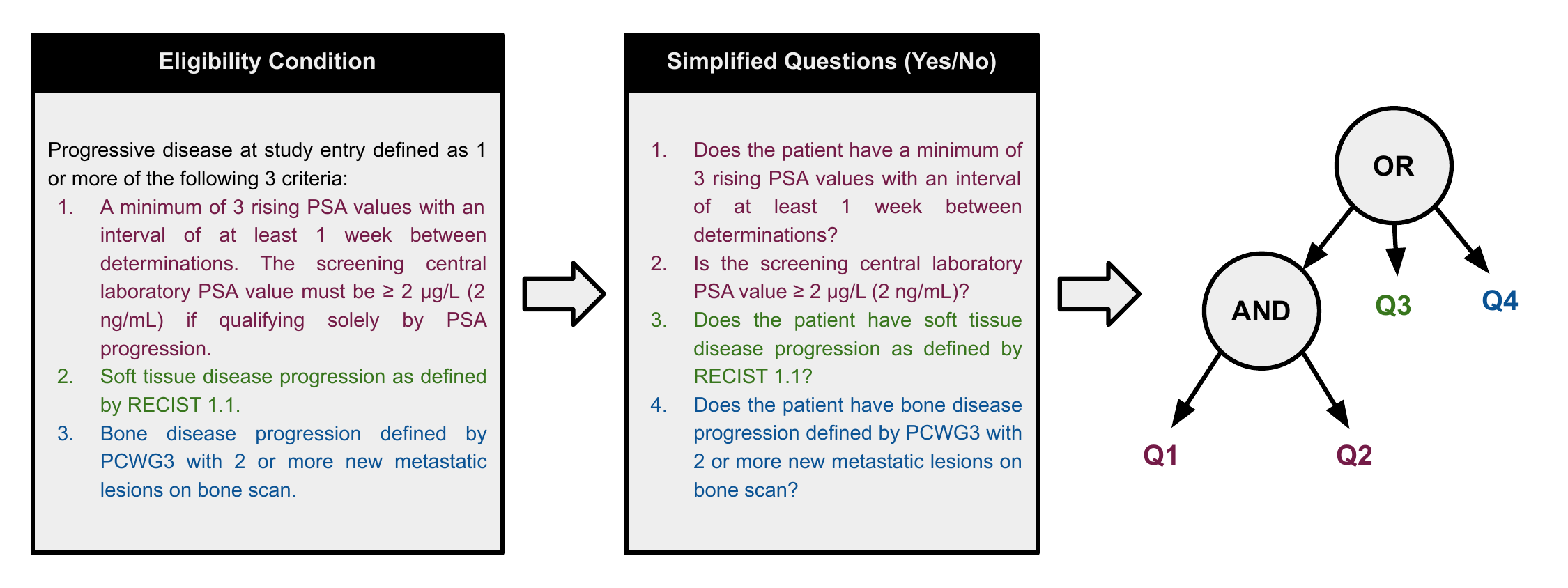}
    \caption{ Each criterion in the clinical trial's free text is condensed into one or more questions using a LLM via an appropriate prompt. For criteria leading to multiple questions, they are structured into a Disjunctive Normal Form (DNF). When assessing whether the patient meets that criteria on the basis of the answer to these questions, we employ these logical constraints to determine if the criteria are fulfilled, rather than burdening the LLM with interpreting all complex questions simultaneously. This approach enables the model to address one straightforward question sequentially.}

    \label{fig:dnf}
\end{figure}

 For each clinical trial \( T \), we utilize our \textit{trial-to-questions} module that can be developed with any LLMs. We used GPT-3.5 in this study as trial criteria is public information and GPT-3.5 is cost effective. This module is tasked with converting the textual content of the trial criteria, as extracted from ClinicalTrials.gov, into a set of simplified, independent questions. Each question is designed to be answerable with responses such as 'Yes', 'No', or 'NA'. Given the inherent complexity of these questions, a single criterion from the trial criteria does not always translate directly into a single question. To address this, each criterion is decomposed into multiple questions that are interconnected through Boolean logic, specifically in disjunctive normal form (DNF) (See \textbf{Fig \ref{fig:dnf}}). This transformation ensures that our downstream modules receive one simplified question at a time, facilitating more straightforward processing and evaluation. This is based on the approach used by \cite{scaling}. We conducted a manual evaluation of the output quality of 50 trials based on three metrics: 1) The percentage of incorrectly formed questions; 2) The percentage of questions missed by the LLM model; and 3) The percentage of incorrectly formed Boolean logic. Only $1\%$ of questions were missed, and merely $2\%$ of questions were incorrectly formed. Additionally, we achieved an accuracy rate of approximately $89\%$ in correctly forming boolean logic at the criterion level.

\subsubsection{Chunking and Retrieval Module}
Each patient's EHRs usually include hundreds of notes, which often surpass the context length capabilities of not only most open-source LLMs but also those of proprietary LLMs. This challenge requires the use of a semantic retrieval engine to manage and extract relevant information from each criterion
\cite {OncoRetriever}. We used the Python spaCy\cite{spacy2} tokenizer, configured with a one-sentence overlap, to chunk the patient notes efficiently. For the retrieval process, we utilized Azure OpenAI's Ada embedding-based model and leveraged cosine similarity between the embeddings of the chunks and the queries derived from the inclusion and exclusion criteria of trials.

\subsubsection{Question-Answering Module}
This module utilizes the information retrieved to answer questions relevant to the clinical trial criteria. Unlike traditional Retrieval-Augmented Generation 
(RAG) pipelines, which typically arrange information chunks based on their relevance scores, our approach organizes the chunks chronologically. This sequence begins with patient age and the date of enrollment. Each chunk is further supplemented with the date of the note from which it was extracted and the type of note. We observed that this approach benefited the model by enabling it to interpret the patient's journey chronologically, which enhanced its ability to answer temporal questions accurately.

We use zero-shot prompting and maintain a temperature setting of zero throughout our experiments to ensure deterministic outputs. We also limit the maximum response length to 8K characters. For experiments utilizing GPT architectures, we employ Azure's HIPAA-compliant APIs. In other scenarios, we deploy models using the vLLM framework on our HIPAA-compliant cloud environment, powered by 4 A100s with 80GB GPU memory each.

The question-answering process also uses a Chain of Thought (CoT) strategy and generates following key-value pairs in the output JSON.
\begin{itemize}
    \item \textbf{Question Explanation:} The model is prompted to explain the requirements of each question, delineate its strategy for answering, and identify additional information that may be required. This step ensures that the model comprehends and recalls the necessary medical concepts from the clinical trial criteria.
    \item \textbf{Answer Explanation:} The model synthesizes and summarizes the information from the patient's records in a step-by-step manner before concluding whether the patient meets the specified criteria.
    \item \textbf{Answer:} The model provides a definitive response of 'Yes', 'No', or 'N/A', based on the analysis conducted.
    \item \textbf{Confidence:} The model quantifies the confidence of its answers on a scale from 1 to 5, where 1 indicates the least confidence and 5 the highest.
\end{itemize}

\subsection{Answers $\rightarrow$ Criteria Module}
The decision logic for determining whether a criteria is met is based on the outputs from a predefined logical tree (See Fig. \ref{fig:dnf}) for that criteria. Each node in this tree corresponds to a specific question as extracted in the Trial composition module. The response to each question can influence the pathway taken through the decision tree, ultimately determining whether that criteria is met or not. In cases where one or more questions' answers are marked as N/A (Not Answerable) for a given criterion, the decision logic involves marginalizing over possible values for all the questions with N/A answers. Mathematically, this process is described as follows:

\begin{equation}
    P(\text{criteria met} | \text{data}) = \sum_{x \in \text{Possible Answers}} P(\text{criteria met} | X = x) \times P(X = x | \text{data})
\end{equation}

Here, $X$ represents the possible combinations of answers to all the questions with 'N/A' as the answer, and $P(X = x | \text{data})$ is the probability of each combination given the patient data. This probability is $\frac{1}{2^{N}}$, where $N$ is the number of questions with 'N/A' answers. The final determination of whether a criteria is met is based on a threshold model:
\newline

\begin{equation}
    \text{Criteria Met} = 
    \begin{cases} 
    \text{Yes} & \text{if } P(\text{criteria met} | \text{data}) > 0.66 \\
    \text{No} & \text{if } P(\text{criteria met} | \text{data}) < 0.34 \\
    \text{N/A} & \text{otherwise} 
    \end{cases}
\end{equation}
\newline

This approach allows for a probabilistic evaluation of criteria even when we do not know the answers to some questions corresponding to that criteria. This approach also takes care of the cases when it is possible to answer the Boolean logic even when answers to few questions are not available. For instance, if the criteria met expression is $Q1 \& Q2$, and we know the answer for $Q1$ is ’No’, we do not need the answer to $Q2$. The expression should resolve to False regardless of the answer to $Q2$. This approach efficiently accommodates such scenarios and ensures robust evaluation of criteria.

\subsubsection{Scoring Module}\label{sec:scoring_module}

This module implements an intelligent scoring module that can be used to rank clinical trials for a patient or vice-versa. Once we have determined the final answer for each criterion in the trial, we employ three strategies to finalize the answer.
\newline

\noindent\textit{\textbf{Simple Counting}} is a straightforward method where each criterion is evaluated independently and then the total number of fulfilled criteria is counted. The final score is then normalized by the total number of conditions (\textbf{Equation \ref{eq:simple_counting}, Fig. \ref{fig:scoring_method}A}):

\begin{equation}\label{eq:simple_counting}
    \text{SCORE}_{simple} = \frac{\text{Number of Criteria Met}}{\text{Total Number of Criteria}}
\end{equation}
\newline

Two other methods are based on Tier-based criteria categorization. These methods employ a more detailed scoring system. Each eligibility criteria is first categorized into different tiers ($T_1$, $T_2$, $T_3$, and $T_4$) according to their clinical importance for patient-trial matching task (See \textbf{Section S2 of Supplementary}) 
, and finally the scores are computed using two approaches: iterative tier and weighted tier.
\newline

\noindent\textbf{\textit{Iterative Tier}} method applies strict rules, traversing from higher to lower tiers until a condition is violated. The final score reflects the proportion of criteria met before encountering a violation, normalized by the total number of conditions (\textbf{Equation \ref{eq:iterative_scoring}, Fig. \ref{fig:scoring_method}B}):
\newline

\begin{equation}\label{eq:iterative_scoring}
    \text{SCORE}_{iterative} = \frac{\text{Number of Criteria Met Until Violation}}{\text{Total Number of Criteria}}
\end{equation}
\newline

\begin{figure}[t]
    \centering
    \includegraphics[width=1\linewidth]{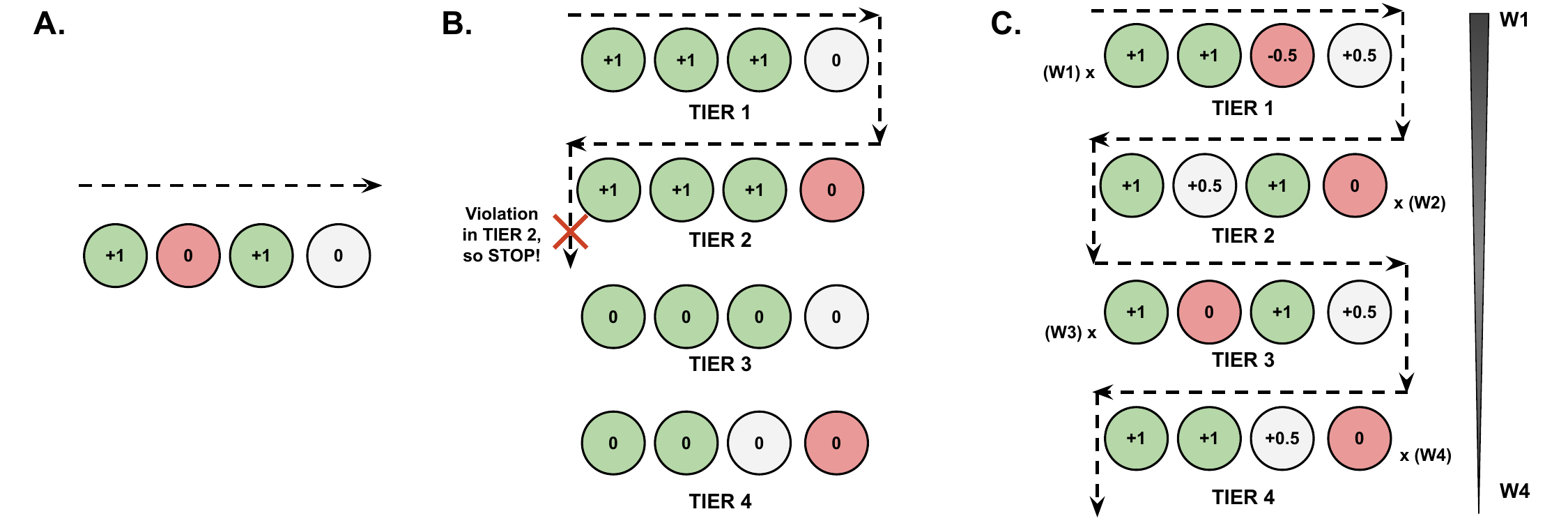}
    \caption{\textbf{Scoring Methods For Patient Trial Matching Score}. \textbf{A. Simple Counting:} In this approach, the total number of criteria met is counted and normalized by the total number of conditions. \textbf{B. Iterative Tier:} Here, we move through all the criteria by categorizing them into tiers. While traversing the nodes, if we encounter a violation (criteria unmet), we stop and use the number of criteria met till that point and normalize it by the total number of conditions. \textbf{C. Weighted Tier:} Similar to iterative tier, this method also utilizes tiers but instead of stopping at the first violation, it continues through the entire tree. It assigns greater importance to the first tier and progressively less weight to lower tiers.}
    \label{fig:scoring_method}
\end{figure}

\noindent\textit{\textbf{Weighted Tier}} method involves ranking criteria by their estimated clinical importance and applying different weights accordingly. It first calculates an intermediate score for each criterion at individual tier level (\textbf{Equation \ref{eq:criteria_level_score}}). The final score calculation is based on weighted averages of the criteria level scores obtained at each tier (\textbf{Equation \ref{eq:weighted_tier_scoring}, Fig. \ref{fig:scoring_method}C}):

\begin{equation}\label{eq:criteria_level_score}
    s(x) = 
    \begin{cases} 
    1 & \text{if } x = 1 \\
    0.5 & \text{if } x = -1 \\
    -0.5 & \text{if $(x = 0)$ and $(tier = T_1)$} \\
    0 & \text{if $(x = 0)$ and $(tier \neq T_1)$}
    \end{cases}
\end{equation}
\newline

\begin{equation}\label{eq:criteria_level_score}
    K = \sum_{k=1}^{4}sgn({|T_k|}) 
\end{equation}
\newline

\begin{equation}\label{eq:weighted_tier_scoring}
    \text{SCORE}_{weighted} = 
    \begin{cases} 
    \frac{1}{K}\sum_{k=1}^{K}{w_k}\left(\frac{\sum_{i \in T_k} s(x_i)}{|T_k|}\right) & \text{if } K \neq 0 \\
     0 & \text{if } K = 0
    \end{cases}
\end{equation}
\newline

where $s(x_i)$ is the score and $x_i$ is the result for the $i^{th}$ criterion, and sgn(x) is the signum/sign function. $T_k$ represents the set of criteria in tier $k$, $w_k$ represents the weight assigned to tier $k$, and $K$ represents the number of tiers with non-zero criteria present in the clinical trial. 

For simplicity we assigned the following weights in this work: $w_1 = 2$, $w_2 = 1.5$, $w_3 = 1$, and $w_4 = 0.5$. However, in the future work, these weights can be tuned to achieve better performance.

\subsection{OncoLLM}
Our model, OncoLLM, is a specialized oncology-focused LLM that has been fine-tuned on a single cancer center's oncology EHR datasets for question-answering tasks, using manually curated outputs. It builds upon the open-source Qwen-1.5 14B model\cite{qwen}, adapting it specifically for this field using both synthetic and real-world data. We chose this model because it was the top performer among similar-sized models on the LLM leaderboard when our experiments began\cite{chiang2024chatbot}. Importantly, no patient data or clinical trials that have been used to report the scores here were added to the training data. The fine-tuning process was carefully chosen to enhance the model's ability to provide medical explanations and reference evidence within a RAG-based framework on EHR records. Annotators crafted ideal responses to EHR-related questions to facilitate the Chain of Thought (CoT) prompting strategy. The model was then trained via supervised fine-tuning on several thousand chunk-question pairs of proprietary data.

\section{Discussion}\label{sec12}
In this study, we developed and validated an end-to-end pipeline for matching patients to clinical trials based on inclusion and exclusion criteria and unstructured patient notes from real-world EHRs. Our findings indicate that leveraging LLMs, with carefully implemented controls, could significantly shift the paradigm for accruing and enrolling eligible patients to clinical trials. Unlike existing processes that rely on time and personnel-intensive manual EHR review, the proposed workflow-based platform can enhance clinical trial efficiency and improve cancer care.
\newline
\newline
\noindent\textbf{Deployment Readiness of the Pipeline:} 
While there are certain limitations, the performance of our proposed pipeline closely matches that of qualified medical professionals in terms of criteria-level accuracy. This significant achievement highlights the models' near readiness for practical deployment. Although the model sometimes makes errors, our citation-based approach—similar to techniques used in several LLM-based search engines—offers a solid pipeline to help humans rectify these inaccuracies.
\newline\newline
\noindent\textbf{Considerations on Model Propriety and Privacy:} Our research challenges the necessity of relying exclusively on proprietary and centralized models in contexts where data privacy is paramount. We demonstrate that smaller models, when fine-tuned appropriately can surpass the performance of their proprietary counterparts. This opens the door to deploying these efficient models in environments where both privacy and cost are of concern. Notably, our model achieves performance metrics comparable to those of GPT4, yet with a significantly lower operational cost, showcasing its potential for scalable applications.
\newline\newline
\noindent\textbf{Feasibility of LLMs in end-to-end Patient-Trial Matching:} To address this question, we extended our pipeline to include a criteria-based ranking system for clinical trials. Our empirical evaluations confirm the feasibility of using LLMs to systematically rank clinical trials against patient records and vice versa. This approach not only facilitated the identification of trials in which patients were actively enrolled but also ended up highlighting trials which the patient was eligible for but was not eventually enrolled. These results substantiate the practicality of employing large language models in the initial phases of clinical trial screening, significantly reducing the time and effort required in trial-patient matching processes.
\newline\newline
\noindent\textbf{Limitations:} While our approach has shown considerable promise, it has limitations that necessitate careful consideration for future development. One of the primary constraints is our reliance solely on unstructured data. Critical information, particularly laboratory values, are often recorded in structured formats and may not be consistently mentioned in patient notes. To address this issue, a hybrid data retrieval approach that integrates both structured and unstructured data could be more effective. Such a model would enhance the comprehensiveness and accuracy of the information retrieval, potentially leading to more precise patient-trial matching. Additionally, our current dependence on embedding-based retrievers presents challenges. Despite their advancements, these retrievers have inherent limitations in processing the complex nuances of medical data. It is imperative to rigorously evaluate the impact of these retrievers on the final outcomes of our pipeline. Although we conducted preliminary evaluations of different retrievers, we did not extend this to the fine-tuning or enhancement of these components.

The accuracy of our end-to-end system, although improved, does not yet meet the ideal standards. Obtaining 'real' ground truth in clinical environments is challenging. We observed significant variations in the responses provided by different annotators, especially for questions where it is difficult to determine whether the available information suffices for accurate criteria assessment. This variation underscores the need for more robust annotation processes and perhaps a reevaluation of the criteria used for determining patient eligibility.
\newline\newline
\noindent\textbf{Directions for Future Research:} These insights pave the way for future work to focus on developing more integrated models that utilize a balanced mix of structured and unstructured data. Enhancing the capabilities and precision of embedding-based retrievers, alongside a more rigorous evaluation framework, will be critical to advancing the technology for clinical application. Further, efforts to standardize the annotation process and refine the accuracy benchmarks will significantly contribute to the reliability of AI-driven clinical trial matching systems.

\backmatter

\bmhead{Supplementary Information}

Comprehensive analyses are provided for each component of our end-to-end pipeline. We conducted experiments on a limited dataset to evaluate the effectiveness of our chunking methods and the distribution of concepts within the inclusion and exclusion criteria of clinical trials. Detailed supplementary materials accompanying this paper include these analyses, offering deeper insights into the procedural nuances and methodological rigor of our study.

\bmhead{Acknowledgements}
We are grateful for the assistance provided by medical professionals, doctors, and research coordinators who shared insights into existing clinical trial matching processes and challenges. We extend special thanks to Harsh Jain, Aniket Jaiswal, and Sebastien Rhodes for their help in interpreting results and offering valuable prompt engineering suggestions. Additionally, we express our gratitude to Sorena Nadaf, Warren Kibbe, and the entire Ci4CC (Cancer Informatics for Cancer Centers) community for facilitating this collaborative effort.
\section*{Declarations}

This study was carried out with the approval of the Institutional Review Board, under PRO \# : 00044894. The authors HS, SG, AB, JT, and MN are employed by Triomics. They were primarily responsible for code implementation, conducting experiments, and various evaluations. NW, AR, AK, and BT from the Medical College of Wisconsin supervised the overall quality and conceptualization of the experiments. They also aided in developing data pipelines, ensuring data quality for dataset preparation, and implementing ranking metrics. YW and TM contributed to conceptualizing experimental methods and supported the writing of the manuscript. The opinions and views expressed in this paper are solely those of the authors and do not reflect the positions or policies of their respective institutions.

\begin{appendices}

\end{appendices}

\bibliography{sn-bibliography}

\pagebreak

\section*{Supplementary Material}
\label{supplementary}

\noindent \textbf{Index:}
\begin{itemize}
    \item \hyperref[supplementary:concordance]{S1 \quad Human Concordance}
    \item \hyperref[supplementary:tier_preparation]{S2 \quad How are the Tiers Prepared?}
    \item \hyperref[supplementary:note_filtering]{S3 \quad Filtering Patient Notes for Experimentation}
    \item \hyperref[supplementary:prompts]{S4 \quad Prompts}
    \item \hyperref[supplementary:sample_output]{S5 \quad Sample Output}
\end{itemize}

\vspace{5mm}

\subsection*{S1 \quad Human Concordance}
\label{supplementary:concordance}

\begin{figure}[H]
    \centering
    \includegraphics[width=1\linewidth]{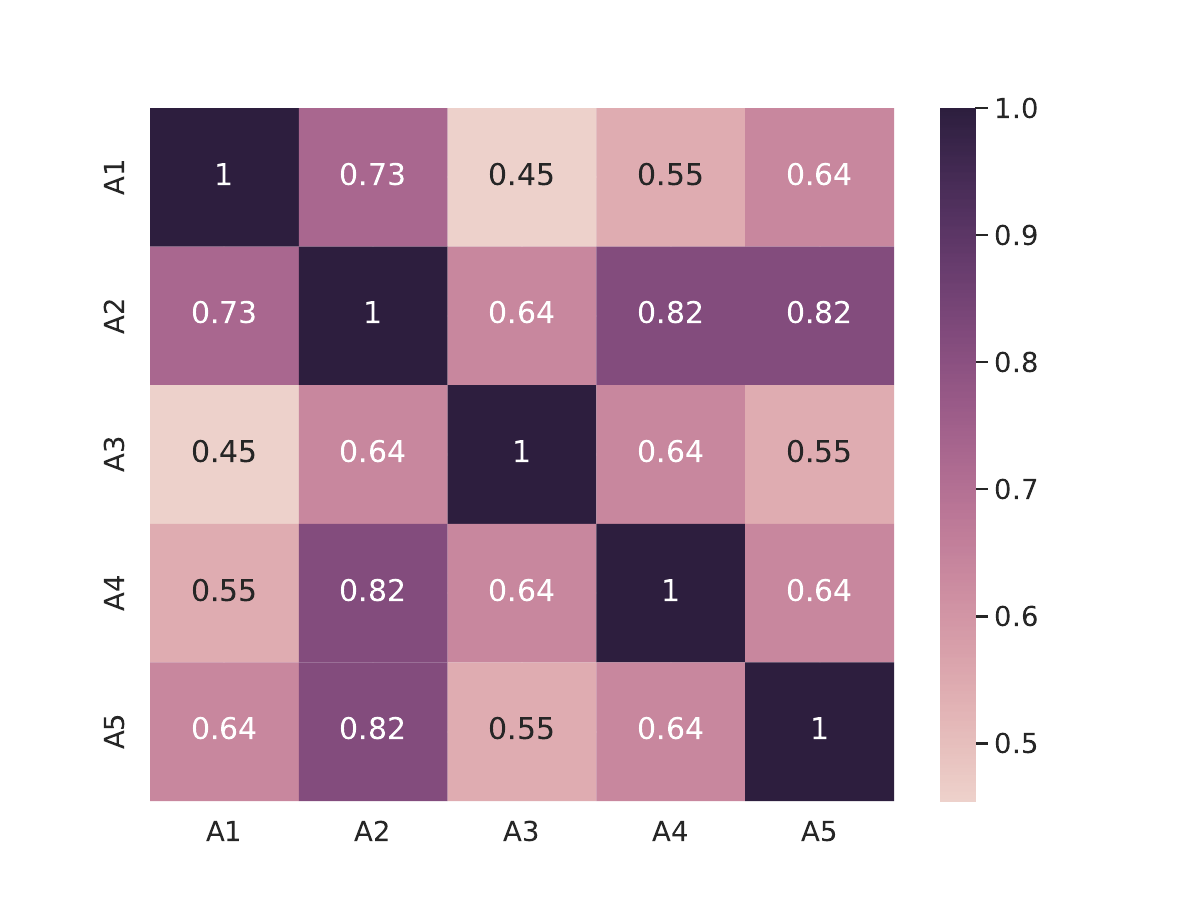}
    \caption{\textbf{Inter Annotator Agreement}. We tasked 5 qualified medical doctors to annotate around 12 questions and utilized their responses to compute their inter-annotator reliability. This matrix illustrates the variability in the rate of disagreement among different annotators, underscoring the task's complexity. We then selected two high-performing annotators from the initial five to annotate 109 additional questions and calculate the final inter-annotator agreement.}
    \label{fig:scoring_method}
\end{figure}

\subsection*{S2 \quad How are the Tiers Prepared?}
\label{supplementary:tier_preparation}

\begin{figure}[H]
    \centering
    \begin{tabular}{|l|c|c|}
        \hline
        \textbf{Concept} & \textbf{Tier}  \\
        \hline
        Cancer Type     & 1               \\
        \hline
        Cancer Subtype     & 1               \\
        \hline
        Cancer Stage     & 1               \\
        \hline
        Cancer Grade/Histology     & 1               \\
        \hline
        Genetic \& Biologic Markers    & 2               \\
        \hline
        Lab/Imaging Criteria     & 2               \\
        \hline
        Prior treatment/surgery     & 2               \\
        \hline
        Comorbidities     & 3               \\
        \hline
        Functional Status     & 4               \\
        \hline

        Others  & 4 \\
        \hline
        
    \end{tabular}
    \caption{Concept-Tier Mapping}
    \label{tab:concept_tier_mapping_data}

\end{figure}

Every criterion of a trial addresses a specific oncology-related concept. We compile these concepts into a "book of concepts". Subsequently, we use GPT-4 to assign a concept to each trial criterion. Recognizing that not all concepts hold equal significance, we categorize them into tiers based on their importance. This tiered system is vital for developing a robust ranking metric, as the penalty or reward for an answer varies by tier, with higher tiers carrying more weight. We consult with domain experts to establish four distinct tiers and allocate each concept to the appropriate tier.

\begin{figure}[H]
    \centering
    \includegraphics[width=0.59\linewidth]{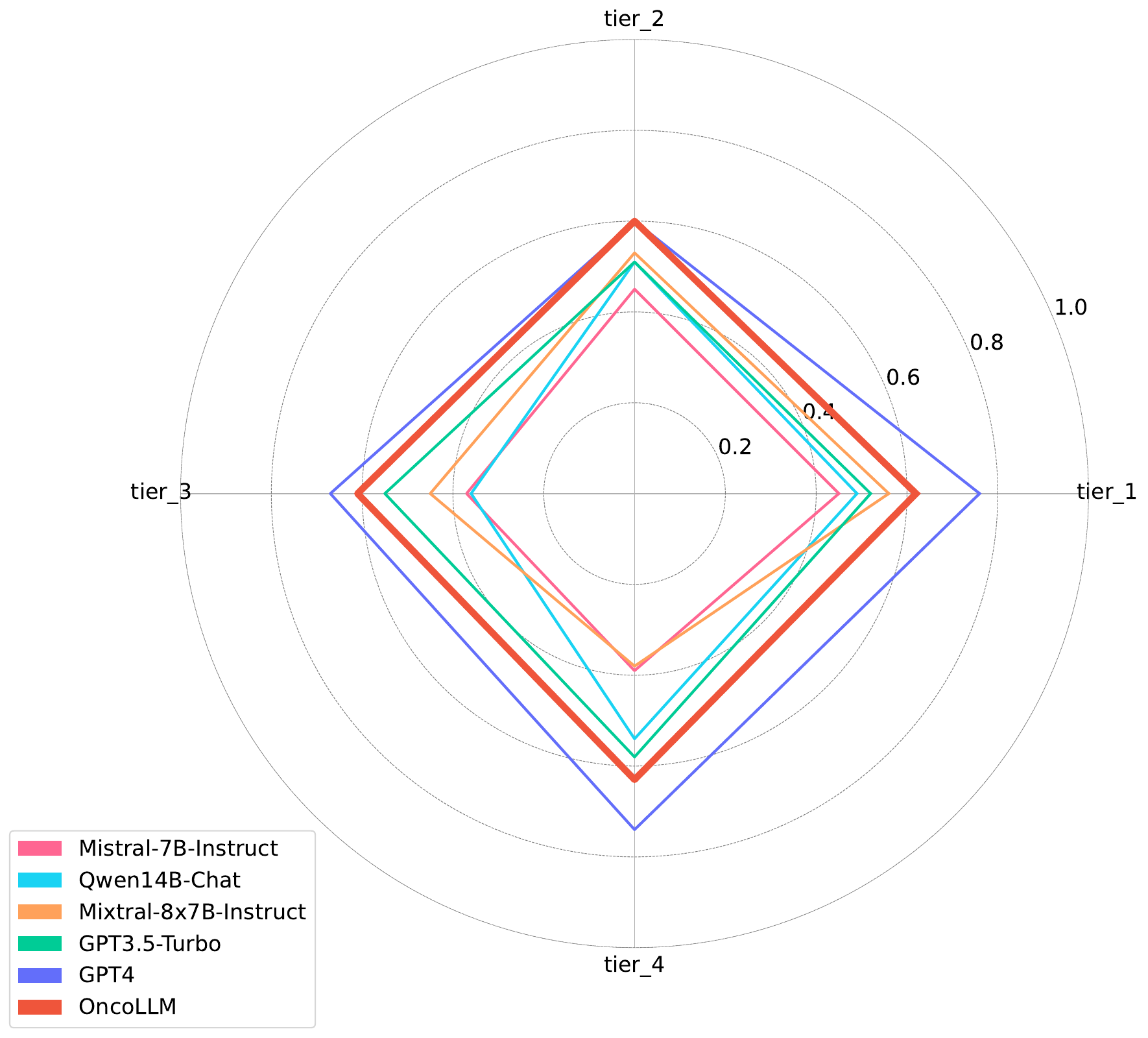}

    \caption{\textbf{Tier-wise Accuracy}}
    \caption{\textbf{ OncoLLM (in red) performs consistent well across all the tiers.}}

    \label{fig:tier_wise_accuracy}
\end{figure}

\vspace{100mm}

\subsection*{S3 \quad Filtering Patient Notes for Experimentation}
\label{supplementary:note_filtering}

Eighty two categories of patients documents were made available to us. After a thorough discourse with domain experts and careful consideration to choose relevant categories, thirteen categories of patient documents were shortlisted and used for experimentation.

\begin{figure}[H]
    \centering
    \begin{tabular}{|l|c|c|}
        \hline
        \textbf{Document Category}  \\
        \hline
        Assessment \& Plan Note                   \\
        \hline
        Brief Op Note               \\
        \hline
        Consults                  \\
        \hline
        Discharge Instructions                \\
        \hline
        Discharge Summary        \\
        \hline
        H\&P                \\
        \hline
        H\&P (View-Only)              \\
        \hline
        Op Note             \\
        \hline
        OR Surgeon           \\
        \hline
        Procedures  \\
        \hline
        Progress Notes  \\
        \hline
        Rad Onc Simulation  \\
        \hline
        Rad Onc Weekly Review  \\
        \hline
    \end{tabular}
    \caption{Selected Categories for Experimentation}
    \label{tab:concept_tier_mapping_data}
\end{figure}

\pagebreak

\end{document}